# Classification of Cross-cultural News Events

Abdul Sittar*
abdul.sittar@ijs.si
Jožef Stefan Institute and Jožef Stefan International
Postgraduate School
Jamova cesta 39
Ljubljana, Slovenia

Dunja Mladenić
dunja.mladenic@ijs.si
Jožef Stefan Institute and Jožef Stefan International
Postgraduate School
Jamova cesta 39
Ljubljana, Slovenia

## ABSTRACT
We present a methodology to support the analysis of culture from text such as news events and demonstrate its usefulness on categorising news events from different categories (society, business, health, recreation, science, shopping, sports, arts, computers, games and home) across different geographical locations (different places in 117 countries). We group countries based on the culture that they follow and then filter the news events based on their content category. The news events are automatically labelled with the help of Hofstede's cultural dimensions. We present combinations of events across different categories and check the performances of different classification methods. We also presents experimental comparison of different number of features in order to find a suitable set to represent the culture.

## KEYWORDS
cultural barrier, news events, text classification

## 1 INTRODUCTION

Culture is defined as a collective programming of the mind which distinguishes the members of one group or category of people from another [9]. It has a huge impact on the lives of people and in result it influences events that involve cross-cultural stakeholders. News spreading is one of the most effective mechanisms for spreading information across the borders. The news to be spread wider cross multiple barriers such as linguistic, economic, geographical, political, time zone, and cultural barriers. Due to rapidly growing number of events with significant international impact, cross-cultural analytics gain increased importance for professionals and researchers in many disciplines, including digital humanities, media studies, and journalism. The most recent examples of such events include COVID-19 and Brexit [1]. There are few determinants that have significant influence on the process of information selection, analysis and propagation. These include cultural values and differences, economic conditions and association between countries. For instance, if two countries are culturally more similar, there are more chances that there will be a heavier news flow between them [10], [3]. In this paper, we focus on classification of news events across different cultures. We select some of the most read daily newspapers and collect information using Event Registry about the news they have published. Event Registry is a system which analyzes news articles, identifies groups of articles that describe the same event and represent them as a single event [7]. The description of the meta data of an event is shown in the Table 1. The main scientific contributions of this paper are the following:

(1) A novel perspective of aligning news events across different cultures through categorising countries and news events.
(2) A cross-cultural automatically annotated dataset in several different domains (Business, Science, Sports, Health etc.).
(3) Experimental comparison of several classification models adopting different set of features (character ngrams, GLOVE embeddings and word ngrams).

**Table 1: The description of the meta data of an event.**

| Attributes | Description |
|---|---|
| title | title of the event |
| summary | summary of the event |
| source | event reported by a news source |
| categories | list of DMOZ categories |
| location | location of the event |

## 2 RELATED WORK

In this section, we review the related literature about the influence of culture, its representation and classification in different fields.

Countries that share a common culture are expected to have heavier news flows between them when reporting on similar events [10]. There are many quantitative studies that found demographic, psychological, socio-cultural, source, system, and content-related aspects [2].

Cross-cultural research and understanding the cultural influences in different fields have competitive advantages. The goal of researching the impact of culture might be to draw conclusions in which way the cultural factors influence a specific corporate action. There are many type of cultures such as societal, organizational, and business culture etc [8].

The hidden nature of cultural behavior causes some difficulties in measurement and defining these. To cope with difficulties, researchers have developed measurements that measure culture on a general scale to compare differences among cultures and management styles. These results can be used to find similarities within a region and differences to other regions. There are many models that have tried to explain cultural differences between societies. Hofstede's national culture dimensions (HNCD) have been widely used and cited in different disciplines [6, 5]. Hofstede's dimensions are the result of a factor analysis at the level of country means of comprehensive survey instrument, aimed at identifying systematic differences in national cultural. Their purpose is to measure culture in countries, societies, sub-groups, and organizations; they are not meant to be regarded as psychological traits.

There is a plethora of research studies that were conducted to understand the cultural influences such as cross-culture privacy and attitude prediction, and cultural influences on today's business. [4] explores how culture affects the technological, organizational, and environmental determinants of machine learning adoption by conducting a comparative case study between Germany and US. Rather than looking at the influence of cultural differences within one domain, we intend to understand association between news events belonging to different domains (society, business, health, recreation, science, shopping, sports, arts, computers, games and home) and different cultures (117 countries from all the continents). We conduct this research to find an appropriate representation and classification of culture across different domains.

## 3 DATA DESCRIPTION
### 3.1 Dataset Statistics

We choose the top 10 daily read newspapers in the world in 2020 [1] and collect the events reported by these newspapers using Event Registry [7] over the time period of 2016-2020. Approximately 8000 events belongs to each newspaper with exception of "Zaman" that has only 900 events. Figure 1 shows the number of events reported by the selected newspapers on a yearly basis. This dataset can be found on the Zenodo repository (version 1.0.0) [2]

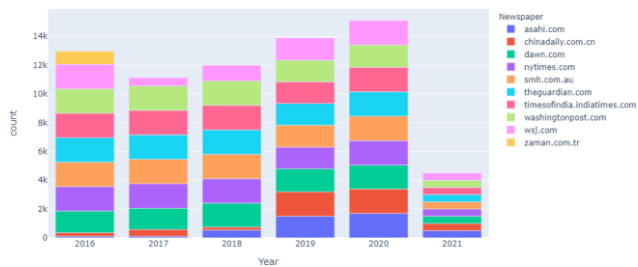

**Figure 1: Each color in a bar represents the total number of events per year by a daily newspaper and a complete bar shows the total number of events per year by all the newspapers.**

The attributes of an event with description are displayed in Table 1. Few attributes are self-explanatory such as title, summary, date, and source. DMOZ-categories are used to represent topics of the content. The DMOZ project is a hierarchical collection of web page links organized by subject matters [3]. Event Registry use top 3 levels of DMoz taxonomy which amount to about 50,000 categories [4].

## 4 MATERIAL AND METHODS
### 4.1 Problem Definition

There are two main parts of the problem that we are addressing. The first part is to label the examples by assigning a culture C to a news event E using its location L. The second part is a multi-class classification task where we predict the culture C of a news event E using its summary description S and its content category G as

---
[1] https://www.trendrr.net/
[2] https://zenodo.org/record/5225053
[3] https://dmoz-odp.org/
[4] https://eventregistry.org/documentation?tab=terminology

provided by the Event Registry. This task can be formulated as:

$$C = f(S, G)$$

C donates the culture of the news event, f is the learning function, S donates summary of a news event and G donates category of a news event (see Table 1).

### 4.2 Methodology

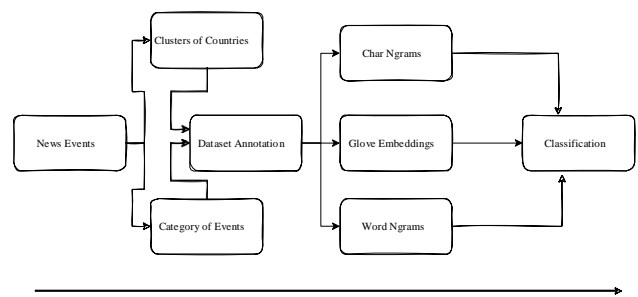

**Figure 2: Classification of cross-cultural news events.**

*4.2.1 Data labeling.* Each news event has information about the type of categories to which it belongs and the location where it happened (see Table 1). Each event has many categories and each category has a weight reflecting its relevance for the event. We only keep the most relevant categories and group the news events based on their categories. For each group of events, we estimate the cultural characteristic of each event through the country of the place where the event occurred. We cluster the countries based on their culture. We utilize the Hofstede's national culture dimensions (HNCD) to represent the culture of a country. We take average of cultural dimensions and call it average cultural score. Based on this score, we find optimal number of clusters using popular clustering algorithm k-means (see Figure 4). Finally, we label each news event with one of the six cultural clusters.

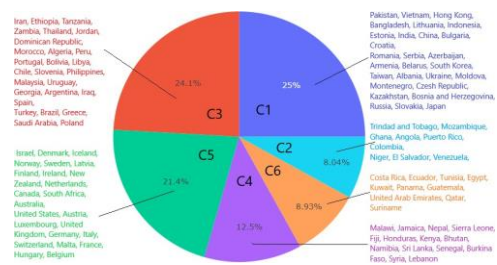

**Figure 3: The pie chart depicts the percentage of the news events that occurred in six different clusters (each cluster consists of a list of countries with similar culture).**

*4.2.2 Data representation.* Each news event in Event Registry has associated categories with it along with a weight (see Table 1), we take the top categories based on their weight. In case of multiple categories with equal weight, we sort them alphabetically and keep the first one. We represent each news event by a short summary S and a set of content categories G.

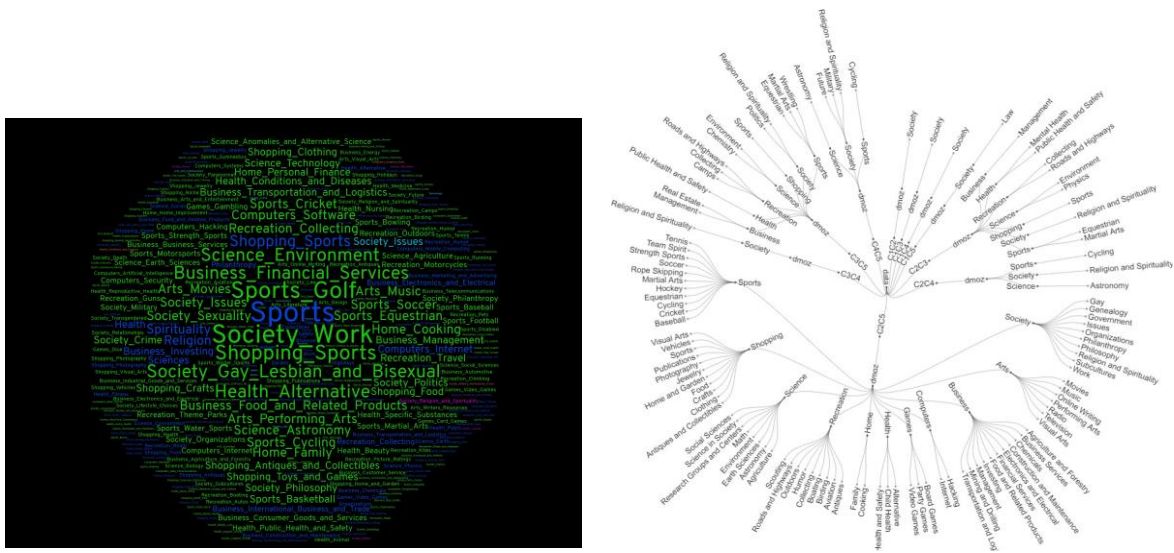

**Figure 4: In word cloud, the color of each word shows cluster to whom it belongs (see Figure 3). Radial dendrograms illustrate the shared categories of news events between the pair of six clusters.**

*4.2.3 Data Modeling.* For multi-class classification task, we use simple classification models (SVM, Decision Tree, KNN, Naive Bayes, Logistic Regression) as well as neural network. For simple classification models, we input character and word ngrams varying the number of ngrams and compare the results. We also use pre-trained Glove embeddings.

## 5 EXPERIMENTAL EVALUATION
### 5.1 Evaluation Metric
For multi-class classification task, we use following most commonly used evaluation measures: accuracy, precision, recall, and F1 score.

## 6 RESULTS AND ANALYSIS
### 6.1 Annotation Results
The results of annotation are six clusters where almost 50% news events belong to the two clusters (shown with red and blue colors) and remaining 50% belong to the other four clusters 3. Looking in each group, we find that clusters do not lies in a specific geographic area or a continent. Rather all the countries in a cluster belong to the different continents. Similarly, these clusters do not have all the countries that are economically rich or poor.

There are more categories in green and red colors in the word cloud (see Figure 4) which represent to the cluster with that colors. Radial dendrograms in Figure 4 present the shared categories between the clusters. In the figure, root of the tree is data and then there are ten pair of clusters that share the same categories. The objective of this whole process was to keep news events according to the category to whom they belongs. Moreover, we can only observe the cultural differences when we have same type of news events from different places.

### 6.2 Classification Results
Fro the experimental results we can see that the best performance is achieved by Logistic Regression, kNN and Decision Tree. The performance of SVM varies depending on the number of selected features: the highest F1-score is achieved with the top 10K or 20K word ngrams using 1 to 3 word ngrams (see Figure 5). Looking at the character ngrams, the highest F1-score is achieved when we select the top 15K characters for all the tested algorithms except Naive Bayes which declines in performance with the growing set of features. Based on these settings, we achieve the highest accuracy (0.85) using Logistic Regression. Using Glove embeddings, we experiment with and without using the category of event. The highest F1-score with and without the category is 0.80 and 0.79 respectively.

## 7 CONCLUSIONS AND FUTURE WORK
For researchers and professionals, it is very important to analyse the cross-cultural differences in different disciplines. As the international impact is increasing and international events are becoming popular, the need to develop some automatic methods is significantly increasing and leaving a blank space. We conducted experiments on news events related to different fields to have a broader look on data and machine learning methods. Further research would be helpful in examining the impact of specific socio-cultural factors on news events. In this research work, we estimate the culture of a specific place by its country, use basic features and simple classification models. To continue this work further, we would like to improve feature set such as by including part of speech tagging (POS) as well as other state of the art embeddings.

## ACKNOWLEDGMENTS
The research described in this paper was supported by the Slovenian research agency under the project J2-1736 Causalify and by the European Union's Horizon 2020 research and innovation programme under the Marie Skłodowska-Curie grant agreement No 812997.

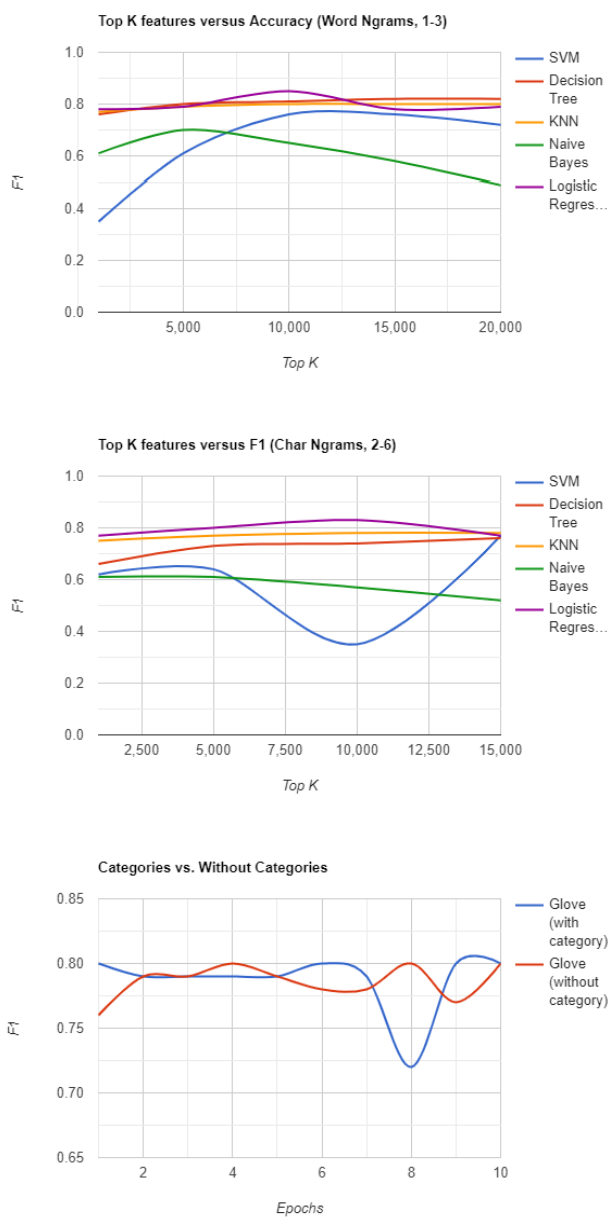

**Figure 5: First two line charts illustrate the variations in F1 score by simple classification models after varying the number of features. The first line chart depicts the results of word ngrams whereas the second one shows the results for character ngrams. The last line graph presents comparison between Glove embeddings (with and without category feature).**